\documentclass{article}
\usepackage{spconf,amsmath,amssymb,graphicx,array,booktabs,cite}
\usepackage{tikz,pgfplots}
\usepackage{colortbl}
\usepackage[hidelinks]{hyperref}
\usepackage{url,balance}
\graphicspath{{Figures/RGBD/}}
\pgfplotsset{compat=1.18}
\title{LVS: Local View Synthesis from Relative Camera Pose by Reusing Previous Views }
\name{
Qizhou Huo$^{1,2}$,
Xuan Sun$^{1,2}$,
Yongfei Guo$^{1,2}$,
Zhipeng Wang$^{1,3}$,
Yuanhao Gong$^{1,3}$
\thanks{This work was supported in part by the National Natural Science Foundation of China under Grant 12471502, the CAS Hundred Talents Program, and the Science and Technology Development Plan Project of Jilin Province, China under Grant 20260204053YY.}
}

\address{
\textit{$^{1}$Changchun Institute of Optics, Fine Mechanics and Physics, Chinese Academy of Sciences}\\
\textit{$^{2}$University of Chinese Academy of Sciences, Beijing, China}\\
\textit{$^{3}$Chinese Academy of Sciences, Beijing, China}
}

\begin{document}
\maketitle

\begin{abstract}
Interactive scene exploration requires frequent view updates, although small camera motions preserve much of the visible content. Conventional 3D Gaussian Splatting nevertheless renders each target view, leaving this image overlap unexploited. Reusing rendered images offers an alternative, but geometric warping alone cannot recover newly exposed content and remains sensitive to depth errors. We propose a per-scene framework that replaces repeated scene rendering for nearby views with relative-pose-guided RGB-D image reuse. Geometric warping uses depth and relative pose to transport source content, while a lightweight multiscale network predicts RGB residuals to correct artifacts and infer missing appearance. Cached source features further reduce repeated computation. On GS-render, residual refinement improves PSNR by 0.72~dB over pure warping; evaluations on captured and rendered scenes demonstrate low query latency. This separation of scene rendering from local view updates supports responsive scene exploration, with potential applications in augmented and virtual reality.
\end{abstract}

\section{Introduction}
Novel view synthesis enables interactive exploration of reconstructed scenes. Conventional scene renderers generate an image for each requested camera pose. During local navigation, however, small camera motions often leave substantial visible content shared with a previously rendered image. This overlap creates an opportunity to reuse the source image, which already contains much of the appearance needed by a nearby target. The central question is how to exploit this overlap while accounting for the image changes introduced by camera motion.

\begin{figure}[t]
\centering
\includegraphics[width=\columnwidth]{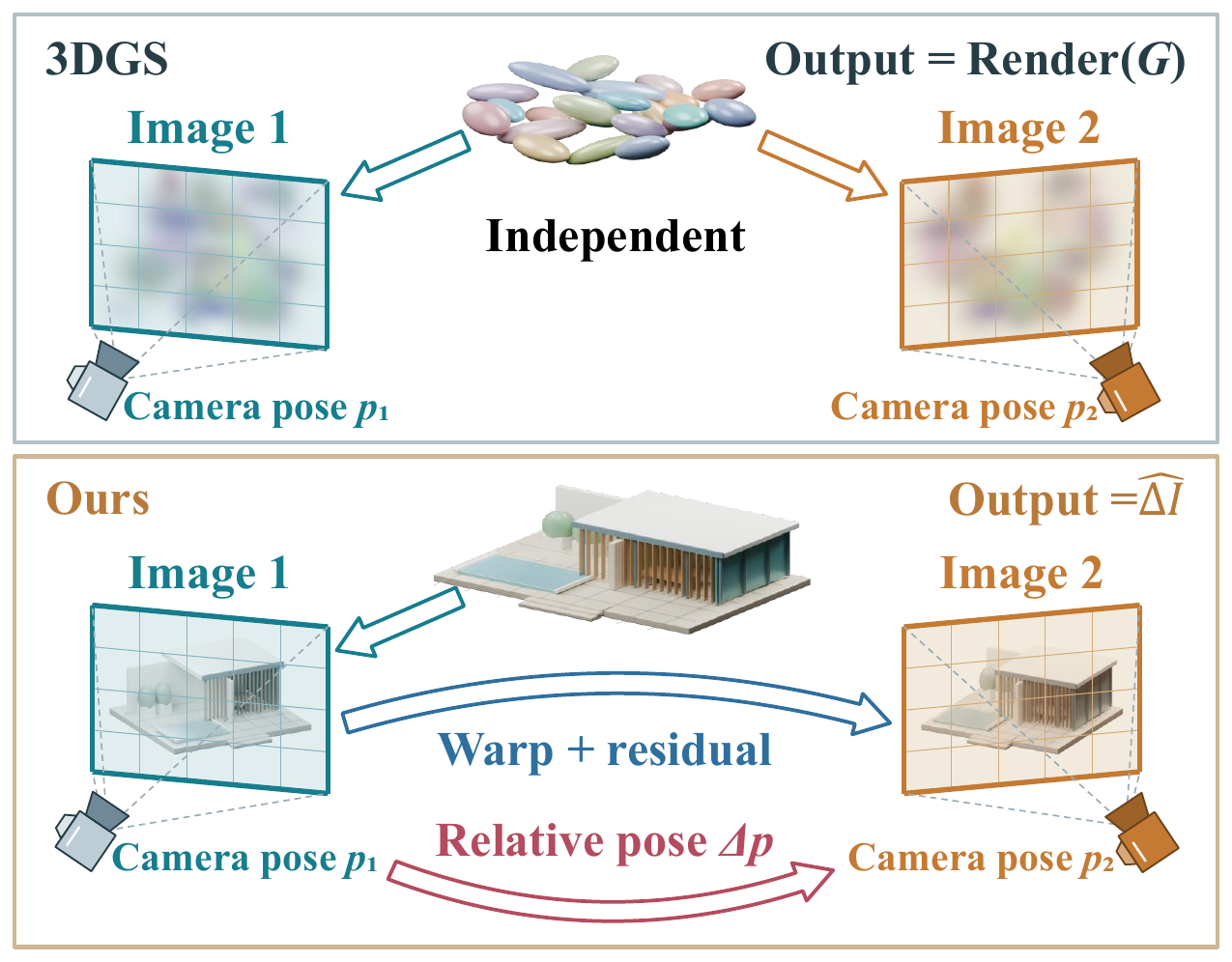}\par
\caption{Local view synthesis. 3DGS renders each target view from the scene representation (top). Our method reuses a rendered RGB-D image through relative-pose-guided warping and RGB residual correction (bottom).}\label{fig:task}
\end{figure}

\begin{figure*}[!t]
\centering
\includegraphics[width=.96\textwidth]{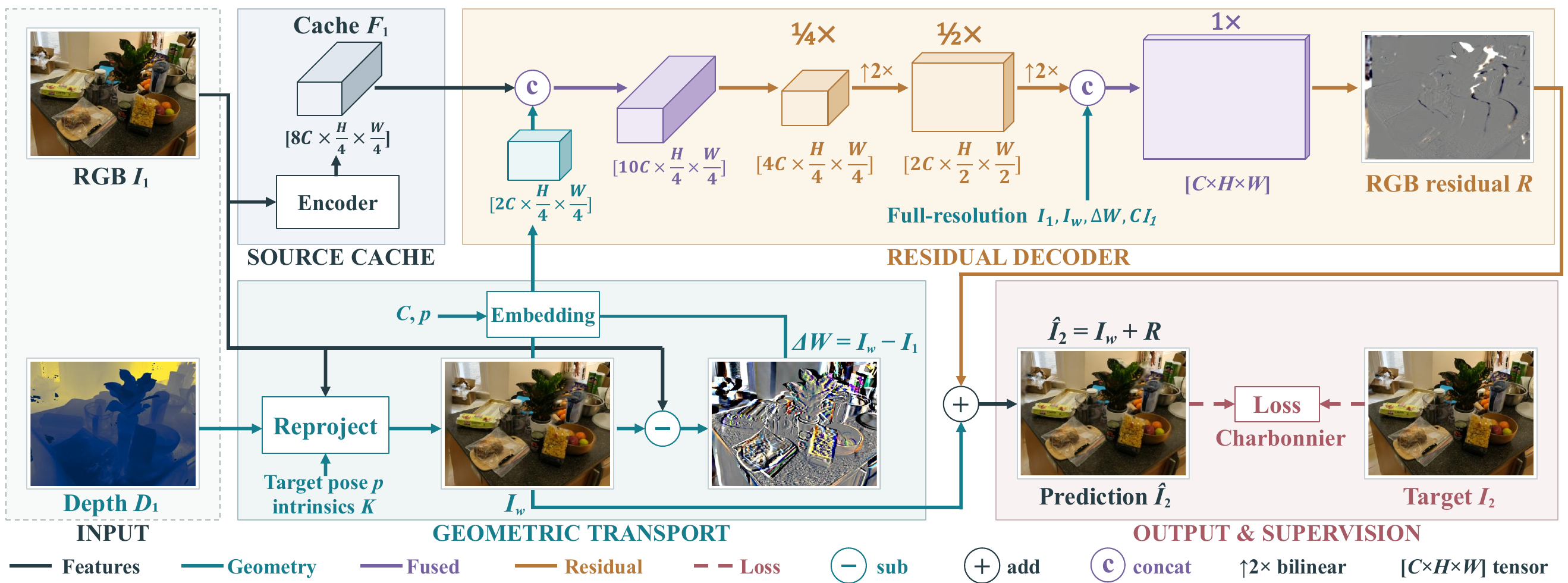}\par
\caption{Framework overview. Depth and relative pose guide warping, while a multiscale decoder combines cached source features and geometric cues to predict an RGB residual. Adding this residual to the warped image yields the target prediction. Dashed red arrows indicate training supervision.}\label{fig:method}
\end{figure*}

\subsection{Scene Rendering and Image Reuse}
Neural Radiance Fields (NeRF) integrate radiance along camera rays, whereas 3D Gaussian Splatting (3DGS) projects and composites Gaussian primitives~\cite{mildenhall2020nerf,kerbl2023gaussians}. Both reuse a persistent scene representation, but conventional target queries still require a new rendering pass. Image-based approaches exploit source observations through learned transformations~\cite{zhou2016appearanceflow,park2017transformation,chen2019continuous}, multiplane representations~\cite{zhou2018stereo,tucker2020singleview,han2022adampi,wang2023lolep}, projected features~\cite{wiles2020synsin,yu2021pixelnerf,wang2021ibrnet}, or learned multiview blending~\cite{hedman2018deepblending}. We focus on synthesizing nearby views from an available rendered RGB-D image.

\subsection{Pose-Guided Warping and Residual Refinement}
Given depth, intrinsics, and relative pose, geometric warping establishes correspondences and transports observed structure~\cite{hou2021depthguided}. However, newly exposed regions lack source observations, while depth inaccuracies, resampling, and view-dependent appearance introduce further errors. CheapNVS learns flow, masks, and inpainting for RGB-D synthesis~\cite{georgiadis2025cheapnvs}; NeuralPassthrough combines depth-based warping with two-view fusion~\cite{xiao2022neuralpassthrough}. Layered inpainting completes hidden content for single-image view synthesis~\cite{shih2020photography,jampani2021slide}. Reusing a single rendered image therefore requires preserving transported detail while correcting regions where correspondence is unreliable or absent.

\subsection{Motivations and Contributions}
We address this problem by separating geometric transport from residual appearance correction (Fig.~\ref{fig:task}). For each static scene, we train a model that refines a pose-guided warp through multiscale RGB residual prediction. Image and geometry guidance retain spatial detail, while cached source features support repeated target queries without invoking the scene renderer. Our main contributions are:
\begin{itemize}
\setlength{\itemsep}{1pt}
\setlength{\parsep}{0pt}
\item We present a relative-pose-guided framework for local view synthesis by reusing rendered RGB-D images.
\item We design a multiscale residual network with source feature reuse and full-resolution RGB/geometry cues.
\item We assess reconstruction quality, query cost, and viewpoint range on real and rendered scenes, with component ablations on rendered data.
\end{itemize}
\begin{figure*}[!t]\centering\includegraphics[width=.92\textwidth]{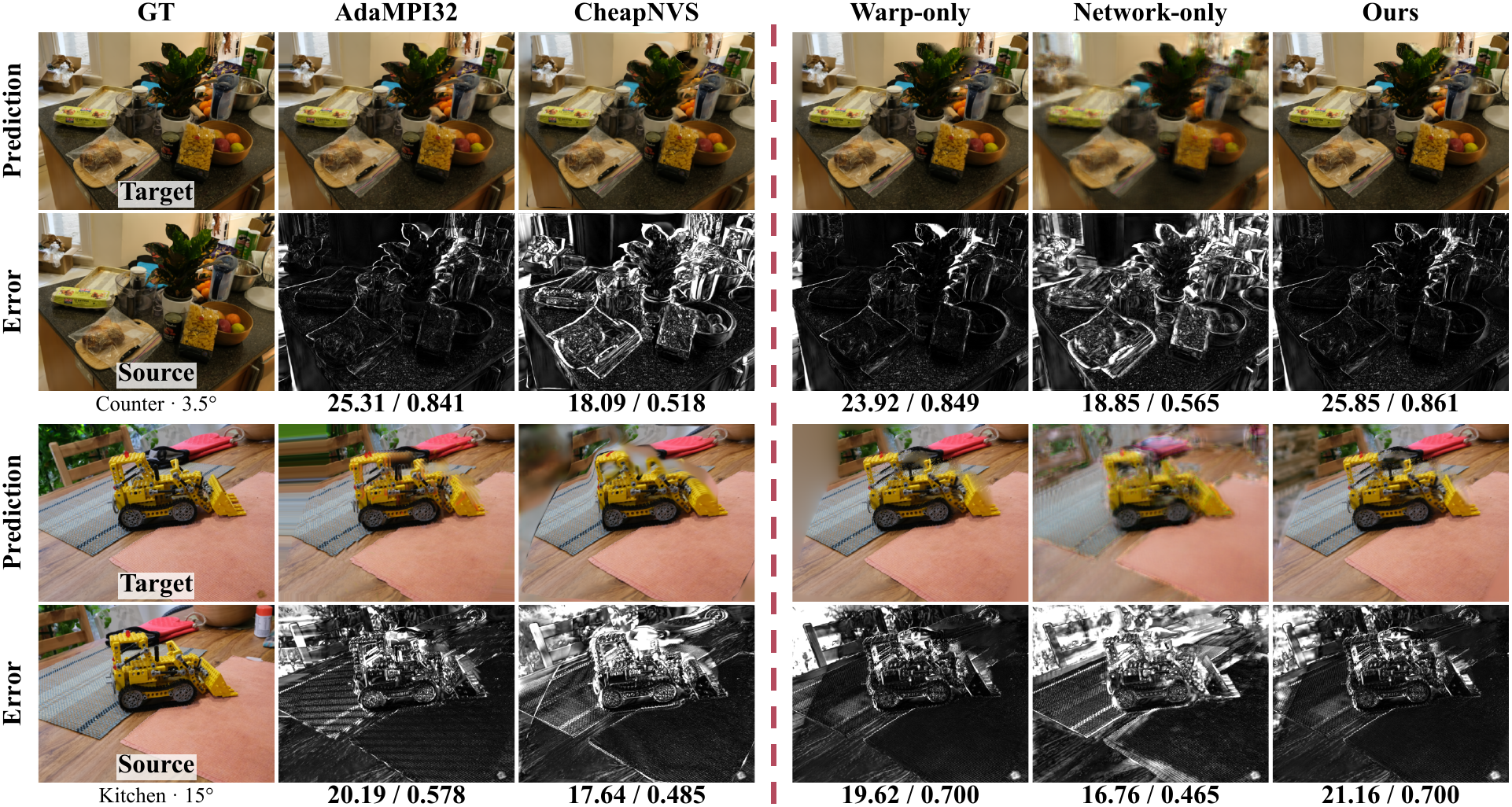}\par
\caption[Qualitative comparison]{Qualitative comparison on Counter ($3.5^\circ$) and Kitchen ($15^\circ$). The first column shows target/source images; other columns show predictions and mean absolute RGB error maps. Scores are PSNR/SSIM. The dashed divider separates external baselines from our component comparisons.}\label{fig:qual}
\end{figure*}

\section{Relative-Pose-Guided Image Reuse with Residual Refinement}
Figure~\ref{fig:method} shows our image reuse pipeline: relative pose guides geometric warping, and a multiscale decoder predicts an RGB correction from source features and geometric cues. The following subsections describe reprojection, residual prediction, and training with source reuse.

\subsection{Relative-Pose-Guided Reprojection}\label{sec:reprojection}
Let $I_1,D_1$ denote the source RGB image and depth, and $K$ the shared camera intrinsic matrix. Rotation matrix $Q$ and translation vector $t$ map source-camera points to target-camera coordinates. A source pixel $x$ projects to target pixel $y$ as
\begin{equation}
\widetilde y\sim K\big[Q\big(\bar D_1(x)K^{-1}\widetilde x\big)+t\big].
\label{eq:projection}
\end{equation}
Here tildes denote homogeneous pixel coordinates with final component one, $\sim$ denotes equality up to scale, and $\bar D_1$ is depth after filling missing values and fitting local inverse-depth planes. For network conditioning, $p$ is the six-dimensional twist (logarithmic coordinates) of the inverse, target-to-source motion~\cite{sola2018micro}.

Source colors are bilinearly splatted and filtered by depth; accepted colors are averaged with their splat weights. Coverage $C$ stores accumulated weights clipped to $[0,1]$, with projected centers set to one. Coverage-weighted coarse-to-fine filling yields the warped image $I_w=\mathcal W(I_1,D_1,K,p)$, where $\mathcal W$ denotes this fixed geometric procedure. The original $C$ is preserved to distinguish filled regions from supported observations. Without projected support, the warp returns $I_1$.

\subsection{Multiscale Residual Refinement}\label{sec:refinement}
The learned encoder $\operatorname{Enc}$ extracts reusable source features $F_1=\operatorname{Enc}(I_1)$. We define $\Delta W=I_w-I_1$ as an RGB image difference. The decoder $\operatorname{Dec}$ predicts the RGB residual $R$, added to $I_w$ to obtain the target prediction $\widehat I_2$:
\begin{equation}
\begin{aligned}
R&=\operatorname{Dec}(F_1,I_1,I_w,\Delta W,C,p),\\
\widehat I_2&=I_w+R.
\end{aligned}
\label{eq:residual}
\end{equation}
Resized $I_w,\Delta W,C$ and broadcast $p$ are linearly embedded and concatenated with $F_1$ at quarter resolution. Refinement proceeds through half to full resolution, where a skip connection supplies $I_1,I_w,\Delta W,C$ to preserve spatial detail. A single linear RGB head outputs $R$.

\subsection{Training and Source Reuse}\label{sec:training}
For each static scene, the encoder and decoder are jointly trained by minimizing the Charbonnier loss~\cite{charbonnier1997regularization}
\begin{equation}
\mathcal L=\operatorname{mean}\!\left(\sqrt{(\widehat I_2-I_2)^2+\epsilon^2}\right).
\label{eq:loss}
\end{equation}
Here, $\mathcal L$ denotes the training loss, $\widehat I_2$ and $I_2$ are the predicted and ground-truth target RGB images, and $\epsilon=10^{-3}$ is a smoothing constant. Squaring and square-root operations are applied elementwise, and $\operatorname{mean}$ averages over all pixels and RGB channels. Zero-initializing the RGB head makes the initial prediction equal to $I_w$. At inference, source features and prepared geometry are cached once per source; each target requires reprojection, residual decoding, and composition.
\begin{table*}[!t]\centering
\caption{Test quality and query cost. Cache and peak memory are in MiB; 3DGS parameter counts refer to stored Gaussian scalars. Best values are bold.}\label{tab:main}
\begingroup\upshape
\begin{tabular}{llrrrrrrr}\toprule Domain & Method & PSNR $\uparrow$ & SSIM $\uparrow$ & LPIPS $\downarrow$ & Params (M) & Query (ms) & Cache & Peak \\\midrule
	7-Scenes & CheapNVS & 16.77 & 0.601 & 0.505 & 9.917 & 8.88 & \textbf{11.4} & 264.2 \\
	7-Scenes & AdaMPI32 & 20.47 & 0.701 & \textbf{0.378} & 18.946 & 5.60 & 403.8 & 7697.1 \\
	7-Scenes & 3DGS & 18.79 & 0.674 & 0.458 & 45.853 & 2.60 & --- & 518.0 \\
	\rowcolor[gray]{0.92}
	7-Scenes & Ours & \textbf{20.74} & \textbf{0.721} & 0.395 & \textbf{0.110} & \textbf{0.95} & 14.4 & \textbf{95.8} \\
	\midrule
	
	GS-render & CheapNVS & 20.33 & 0.622 & 0.329 & 9.917 & 8.75 & \textbf{11.4} & 264.2 \\
	GS-render & AdaMPI32 & 25.89 & 0.791 & 0.213 & 18.946 & 5.72 & 403.8 & 7697.1 \\
	GS-render & 3DGS & \textbf{36.65} & \textbf{0.974} & \textbf{0.053} & 55.421 & 3.06 & --- & 582.1 \\
	\rowcolor[gray]{0.92}
	GS-render & Ours & 26.89 & 0.841 & 0.189 & \textbf{0.110} & \textbf{0.98} & 14.4 & \textbf{95.8} \\
	\midrule

	Blender & CheapNVS & 26.16 & 0.804 & 0.285 & 9.917 & 9.02 & \textbf{11.4} & 264.2 \\
	Blender & AdaMPI32 & 29.97 & 0.864 & 0.234 & 18.946 & 5.60 & 403.8 & 7697.1 \\
	Blender & 3DGS & 26.90 & 0.846 & 0.385 & 16.427 & 2.06 & --- & 226.3 \\
	
	\rowcolor[gray]{0.92}
	Blender & Ours & \textbf{30.57} & \textbf{0.882} & \textbf{0.179} & \textbf{0.110} & \textbf{0.98} & 14.4 & \textbf{95.8} \\
	\bottomrule\end{tabular}
\endgroup
\end{table*}

\section{Experiments}\label{sec:experiments}
We evaluate rendered-image reuse through reconstruction quality, component ablations, and viewpoint changes. The following experiments assess how geometric transport and residual refinement support local synthesis, together with the computational cost of source reuse.

\subsection{Experimental Setup}\label{sec:setup}
We evaluate rendered RGB-D data from GS-render (three Mip-NeRF 360 scenes~\cite{mip360}) and Blender (Barcelona Pavilion), alongside captured 7-Scenes data~\cite{seven}. Images are $640\times480$, with disjoint source splits and viewpoint changes up to $15^\circ$; rendered tests include held-out motion directions. Our models use per-scene Adam~\cite{kingma2015adam} training ($10^{-3}$) and validation-selected checkpoints.

Baselines are reimplemented CheapNVS~\cite{georgiadis2025cheapnvs}, pretrained AdaMPI32~\cite{han2022adampi}, and scene-trained 3DGS~\cite{kerbl2023gaussians}, with different training regimes and inputs. PSNR, SSIM~\cite{ssim}, and LPIPS~\cite{lpips} use identical held-out targets and equal averaging across sources, angular bins, and scenes. Timing uses an RTX 5090, FP32, batch one, and CUDA synchronization; memory excludes metrics and target RGB.

\subsection{Local View Synthesis Results}\label{sec:results}
Table~\ref{tab:main} shows that Ours achieves the highest PSNR and SSIM among the evaluated methods in each domain. On GS-render, it exceeds AdaMPI32 by 1.00~dB PSNR, supporting reuse of an existing rendering for nearby target views. This advantage also holds on Blender, while 7-Scenes tests the framework with captured RGB-D inputs. AdaMPI32 nevertheless achieves lower LPIPS on 7-Scenes.

Figure~\ref{fig:qual} compares identical source--target pairs. At $3.5^\circ$, residual refinement reduces errors around object boundaries relative to Pure Warp. At $15^\circ$, it improves PSNR while visible distortions remain, illustrating the difficulty of correcting larger viewpoint changes from a single source.
\subsection{Ablation Studies}\label{sec:ablation}
Table~\ref{tab:ablation} evaluates geometry and residual refinement on the three GS-render scenes. Pure Warp returns $I_w$, while Pure Network predicts image change from RGB and pose. Quarter/Half variants predict residuals at reduced resolution and upsample them. Learned variants share optimization and checkpoint selection.
\begin{table}[!t]\centering
\caption{Component ablation on GS-render, averaged over three scenes. Best values are bold.}\label{tab:ablation}
\small\setlength{\tabcolsep}{4pt}\begingroup\upshape
\begin{tabular}{lrrrr}\toprule Variant & PSNR $\uparrow$ & SSIM $\uparrow$ & LPIPS $\downarrow$ & Params (K) \\\midrule
Warp-only & 26.166 & 0.8340 & 0.1898 & \textbf{0.0} \\
Network-only & 19.659 & 0.5833 & 0.4299 & 109.7 \\
Residual-1/4 & 26.526 & 0.8338 & 0.1910 & 104.5 \\
Residual-1/2 & 26.706 & 0.8343 & 0.1902 & 108.7 \\
Residual-Full & \textbf{26.887} & \textbf{0.8407} & \textbf{0.1888} & 110.4 \\
\bottomrule\end{tabular}
\endgroup
\end{table}
Full refinement improves Pure Warp by 0.721~dB PSNR, with a smaller LPIPS change. PSNR increases from quarter to full resolution, supporting progressive refinement. Pure Network performs substantially worse, indicating the value of depth-guided transport in this system. Since capacities and inputs differ, these comparisons assess complete component configurations.

\subsection{Viewpoint Range and Computational Cost}\label{sec:range_cost}
\begin{figure}[!t]
\centering
\includegraphics[width=\columnwidth]{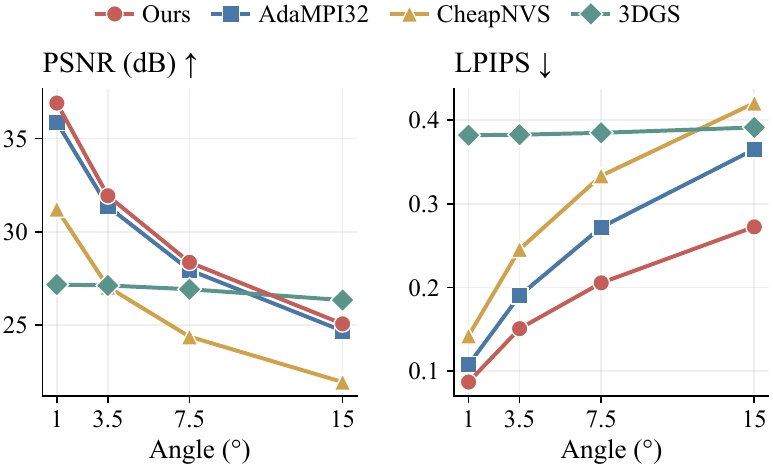}\par
\caption[Viewpoint range]{Viewpoint sensitivity on Blender. PSNR (left) and LPIPS (right) are evaluated at matched target poses.}\label{fig:range}
\end{figure}
Figure~\ref{fig:range} examines viewpoint changes on Blender. Ours has higher PSNR at smaller angles, but 3DGS overtakes it at $15^\circ$; Ours retains lower LPIPS. This crossover illustrates the limited viewpoint range of single-source reuse. Blender covers one scene; cross-scene transfer is not evaluated.

Table~\ref{tab:main} reports query latency of 0.95--0.98~ms for Ours, with a 14.4~MiB source cache and 95.8~MiB peak allocated memory. Source reuse shares feature extraction and geometry preparation across queries; target-dependent reprojection and residual prediction remain necessary for each new view.
\section{Conclusion}
We presented a per-scene framework for local view synthesis by reusing rendered RGB-D images without rerendering the scene. Relative-pose-guided warping preserves shared content, while multiscale residual refinement corrects transport errors and infers missing appearance. Experiments on captured and rendered scenes demonstrate low query latency, and GS-render ablations support the benefit of residual refinement. These findings highlight image reuse as a practical approach to responsive scene navigation, with potential AR/VR applications. Performance declines with larger viewpoint changes; extending the supported range and enabling cross-scene generalization remain future work.
\clearpage
\balance
\bibliographystyle{IEEEbib}
\bibliography{refs}

@inproceedings{mildenhall2020nerf,
  author        = {Mildenhall, Ben and Srinivasan, Pratul P. and Tancik, Matthew and others},
  title         = {{NeRF}: Representing Scenes as Neural Radiance Fields for View Synthesis},
  booktitle     = {Proc. ECCV},
  series        = {Lecture Notes in Computer Science},
  volume        = {12346},
  pages         = {405--421},

  year          = {2020},
  doi           = {10.1007/978-3-030-58452-8_24}
}

@article{kerbl2023gaussians,
  author        = {Kerbl, Bernhard and Kopanas, Georgios and Leimk{\"u}hler, Thomas and others},
  title         = {{3D} {Gaussian} Splatting for Real-Time Radiance Field Rendering},
  journal       = {ACM Trans. Graph.},
  volume        = {42},
  number        = {4},
  articleno     = {139},
  pages         = {1--14},
  year          = {2023},
  doi           = {10.1145/3592433}
}

@inproceedings{park2017transformation,
  author        = {Park, Eunbyung and Yang, Jimei and Yumer, Ersin and others},
  title         = {Transformation-Grounded Image Generation Network for Novel {3D} View Synthesis},
  booktitle     = {Proc. CVPR},
  pages         = {702--711},

  year          = {2017},
  doi           = {10.1109/CVPR.2017.82}
}

@inproceedings{chen2019continuous,
  author        = {Chen, Xu and Song, Jie and Hilliges, Otmar},
  title         = {Monocular Neural Image Based Rendering with Continuous View Control},
  booktitle     = {Proc. ICCV},
  pages         = {4089--4099},

  year          = {2019},
  doi           = {10.1109/ICCV.2019.00419}
}

@inproceedings{tucker2020singleview,
  author        = {Tucker, Richard and Snavely, Noah},
  title         = {Single-View View Synthesis with Multiplane Images},
  booktitle     = {Proc. CVPR},
  pages         = {548--557},
  year          = {2020},
  doi           = {10.1109/CVPR42600.2020.00063}
}

@inproceedings{han2022adampi,
  author        = {Han, Yuxuan and Wang, Ruicheng and Yang, Jiaolong},
  title         = {Single-View View Synthesis in the Wild with Learned Adaptive Multiplane Images},
  booktitle     = {Proc. SIGGRAPH},
  pages         = {1--8},

  year          = {2022},
  doi           = {10.1145/3528233.3530755}
}

@inproceedings{wang2023lolep,
  author        = {Wang, Cong and Wang, Yu-Ping and Manocha, Dinesh},
  title         = {{LoLep}: Single-View View Synthesis with Locally-Learned Planes and Self-Attention Occlusion Inference},
  booktitle     = {Proc. ICCV},
  pages         = {10807--10817},
  year          = {2023},
  doi           = {10.1109/ICCV51070.2023.00995}
}

@inproceedings{wiles2020synsin,
  author        = {Wiles, Olivia and Gkioxari, Georgia and Szeliski, Richard and others},
  title         = {{SynSin}: End-to-End View Synthesis from a Single Image},
  booktitle     = {Proc. CVPR},
  pages         = {7465--7475},

  year          = {2020},
  doi           = {10.1109/CVPR42600.2020.00749}
}

@inproceedings{yu2021pixelnerf,
  author        = {Yu, Alex and Ye, Vickie and Tancik, Matthew and others},
  title         = {{pixelNeRF}: Neural Radiance Fields from One or Few Images},
  booktitle     = {Proc. CVPR},
  pages         = {4576--4585},

  year          = {2021},
  doi           = {10.1109/CVPR46437.2021.00455}
}

@inproceedings{hou2021depthguided,
  author        = {Hou, Yuxin and Solin, Arno and Kannala, Juho},
  title         = {Novel View Synthesis via Depth-Guided Skip Connections},
  booktitle     = {Proc. WACV},
  pages         = {3118--3127},

  year          = {2021},
  doi           = {10.1109/WACV48630.2021.00316}
}

@inproceedings{georgiadis2025cheapnvs,
  author        = {Georgiadis, Konstantinos and Yucel, Mehmet Kerim and Saa-Garriga, Albert},
  title         = {{CheapNVS}: Real-Time On-Device Narrow-Baseline Novel View Synthesis},
  booktitle     = {Proc. ICASSP},
  pages         = {1--5},

  year          = {2025},
  doi           = {10.1109/ICASSP49660.2025.10888972}
}

@inproceedings{xiao2022neuralpassthrough,
  author        = {Xiao, Lei and Nouri, Salah and Hegland, Joel and others},
  title         = {{NeuralPassthrough}: Learned Real-Time View Synthesis for {VR}},
  booktitle     = {Proc. SIGGRAPH},
  pages         = {1--9},

  year          = {2022},
  doi           = {10.1145/3528233.3530701}
}

@misc{sola2018micro, author={Joan Sol{\`a} and J{\'e}r{\'e}mie Deray and Dinesh Atchuthan}, title={A Micro {Lie} Theory for State Estimation in Robotics}, howpublished={arXiv:1812.01537}, year={2018}}

@inproceedings{seven,author={Jamie Shotton and Ben Glocker and Christopher Zach and others},title={Scene Coordinate Regression Forests for Camera Relocalization in {RGB-D} Images},booktitle={Proc. CVPR},year={2013},pages={2930--2937}}

@inproceedings{mip360,author={Jonathan T. Barron and Ben Mildenhall and Dor Verbin and others},title={{Mip-NeRF 360}: Unbounded Anti-Aliased Neural Radiance Fields},booktitle={Proc. CVPR},year={2022},pages={5470--5479}}

@article{ssim, author={Zhou Wang and Alan C. Bovik and Hamid R. Sheikh and others}, title={Image Quality Assessment: From Error Visibility to Structural Similarity}, journal={IEEE Trans. Image Process.}, volume={13}, number={4}, pages={600--612}, year={2004}, doi={10.1109/TIP.2003.819861}}

@inproceedings{lpips,author={Richard Zhang and Phillip Isola and Alexei A. Efros and others},title={The Unreasonable Effectiveness of Deep Features as a Perceptual Metric},booktitle={Proc. CVPR},year={2018},pages={586--595}}

@inproceedings{zhou2016appearanceflow,
  author        = {Zhou, Tinghui and Tulsiani, Shubham and Sun, Weilun and others},
  title         = {View Synthesis by Appearance Flow},
  booktitle     = {Proc. ECCV},
  series        = {Lecture Notes in Computer Science},
  volume        = {9908},
  pages         = {286--301},

  year          = {2016},
  doi           = {10.1007/978-3-319-46493-0_18}
}

@article{zhou2018stereo,
  author = {Zhou, Tinghui and Tucker, Richard and Flynn, John and others},
  title = {Stereo Magnification: Learning View Synthesis Using Multiplane Images},
  journal = {ACM Trans. Graph.},
  volume = {37},
  number = {4},
  pages = {1--12},
  year = {2018},
  doi = {10.1145/3197517.3201323}
}

@article{hedman2018deepblending,
  author = {Hedman, Peter and Philip, Julien and Price, True and others},
  title = {Deep Blending for Free-Viewpoint Image-Based Rendering},
  journal = {ACM Trans. Graph.},
  volume = {37},
  number = {6},
  pages = {1--15},
  year = {2018},
  doi = {10.1145/3272127.3275084}
}

@inproceedings{shih2020photography,
  author = {Shih, Meng-Li and Su, Shih-Yang and Kopf, Johannes and others},
  title = {{3D} Photography Using Context-Aware Layered Depth Inpainting},
  booktitle = {Proc. CVPR},
  pages = {8028--8038},
  year = {2020},
  doi = {10.1109/CVPR42600.2020.00805},
  url = {https://openaccess.thecvf.com/content_CVPR_2020/html/Shih_3D_Photography_Using_Context-Aware_Layered_Depth_Inpainting_CVPR_2020_paper.html}
}

@inproceedings{jampani2021slide,
  author = {Jampani, Varun and Chang, Huiwen and Sargent, Kyle and others},
  title = {{SLIDE}: Single Image {3D} Photography With Soft Layering and Depth-Aware Inpainting},
  booktitle = {Proc. ICCV},
  pages = {12518--12527},
  year = {2021},
  doi = {10.1109/ICCV48922.2021.01229},
  url = {https://openaccess.thecvf.com/content/ICCV2021/html/Jampani_SLIDE_Single_Image_3D_Photography_With_Soft_Layering_and_Depth-Aware_ICCV_2021_paper.html}
}

@inproceedings{wang2021ibrnet,
  author = {Wang, Qianqian and Wang, Zhicheng and Genova, Kyle and others},
  title = {{IBRNet}: Learning Multi-View Image-Based Rendering},
  booktitle = {Proc. CVPR},
  pages = {4690--4699},
  year = {2021},
  doi = {10.1109/CVPR46437.2021.00466},
  url = {https://openaccess.thecvf.com/content/CVPR2021/html/Wang_IBRNet_Learning_Multi-View_Image-Based_Rendering_CVPR_2021_paper.html}
}

@article{charbonnier1997regularization,
  author = {Charbonnier, P. and Blanc-Feraud, L. and Aubert, G. and Barlaud, M.},
  title = {Deterministic Edge-Preserving Regularization in Computed Imaging},
  journal = {IEEE Trans. Image Process.},
  volume = {6},
  number = {2},
  pages = {298--311},
  year = {1997},
  doi = {10.1109/83.551699}
}

@inproceedings{kingma2015adam,
  author = {Kingma, Diederik P. and Ba, Jimmy},
  title = {{Adam}: A Method for Stochastic Optimization},
  booktitle = {Proc. ICLR},
  year = {2015},
  url = {https://arxiv.org/abs/1412.6980}
}
\end{document}